\documentclass[letterpaper, 10 pt, conference]{ieeeconf}  % Comment this line out if you need a4paper

\IEEEoverridecommandlockouts                              % This command is only needed if 
\usepackage{blindtext}
\usepackage{graphicx}

\usepackage[font=small]{caption}
\usepackage{amssymb}
\usepackage[cmex10]{amsmath}
\usepackage{float}
\usepackage{subcaption}
\usepackage{ tipa }
\usepackage{hyperref}
\usepackage[table]{xcolor}
\usepackage{mathpazo} % Use the Palatino font
\usepackage[final]{pdfpages}
\usepackage{csquotes} % for "quotes"
\usepackage{cite} % for mulicitation like this [1-3]
\usepackage{siunitx} % for SI units
\usepackage{scrextend} % for footnodes 
\usepackage{courier} % for code font (Courier font)
\usepackage{placeins} % for \FloatBarrier
\usepackage{pgffor} % enables for loops. Format: \foreach \index in {1, ..., 8} { code here with use of \index}
\usepackage{csquotes}
\usepackage{tabularx}
\usepackage[english]{babel} % blind text
\usepackage{blindtext}
\usepackage{array} %improves the standard LaTeX2e array and tabular environments to provide better
\usepackage{mdwmath} % improved equation formatting
\usepackage{eqparbox} %same
\renewcommand{\thetable}{\arabic{table}}

\let\labelindent\relax % evoid \labelindent conflict with other packages before using enumitem
\usepackage{enumitem}

\usepackage{multicol} % for multicolon bibliography (list of references)
\usepackage{etoolbox} % for multicolon bibliography (list of references)
\usepackage{hyperref} % link references (not really working for cites)
\usepackage{flushend}
\makeatletter
\let\NAT@parse\undefined
\makeatother

\begin{document}
    \title{Towards Intention Prediction for Handheld Robots: a Case of Simulated Block Copying}
    \author{Janis Stolzenwald and Walterio W. Mayol-Cuevas \thanks{ 
            Department of Computer Science, University of Bristol, UK,
            {\color{white}....} % required for new line hack
            janis.stolzenwald.2015@my.bristol.ac.uk, wmayol@cs.bris.ac.uk
            %Department of Computer Science, University of %Bristol {\color{white}...................................................} 
            %({\tt\small janis.stolzenwald.2015@my.bristol.ac.uk})
        }
    }
    \maketitle
%    \begin{abstract}
%        %\boldmath  
%    \end{abstract}
%    
            \begin{abstract}
            %\boldmath
        
        Within this work, we explore intention inference for user actions in the context of a handheld robot setup. Handheld robots share the shape and properties of handheld tools while being able to process task information and aid manipulation. Here, we propose an intention prediction model to enhance cooperative task solving. Within a block copy task, we collect eye gaze data using a robot-mounted remote eye tracker which is used to create a profile of visual attention for task-relevant objects in the workspace scene. These profiles are used to make predictions about user actions i.e. which block will be picked up next and where it will be placed. Our results show that our proposed model can predict user actions well in advance with an accuracy of 87.94\% (\SI{500}{ms} prior)  for picking and 93.25\% (\SI{1500}{ms} prior) for placing actions. 
       % This will be a really nice abstract. So nice in fact that everyone would immediately buy this paper since logging into the university account just takes too long. But the shininess of this abstract is not everything, it is much more about the amazing content. For example, it will say very short, precisely and complete, what this paper is about without saying too much. At the end of the day, we want people to actually read the paper, right. While introducing the reader about the motivation e.g. how super important the paper is, we also advertise our extraordinarily sophisticated approach and method which was used to completely answer the research question. No doubt about that, when the reader is done with the paper, he or she will know much more about this world-leading research than before.  
        \end{abstract}

    
    %    \begin{IEEEkeywords}
    %        HRI, social robots, gesture, mimicry
    %    \end{IEEEkeywords}

    % \fi
    %
    % For peer review papers, this IEEEtran command inserts a page break and
    \IEEEpeerreviewmaketitle

    \section{Introduction}
    % extended version: we could include that in humans, fluid interaction is realised through vise versa action anticipation.
    A Handheld robot shares properties of a handheld tool while being enhanced with autonomous motion as well as the ability to process task-relevant information and user signals. 
    Earlier work in this new field introduced first prototypes which demonstrate that robotic guiding gestures \cite{GreggSmith:2015bh} as well as on-display visual feedback \cite{GreggSmith:2016hn} lead to a level of cooperative performance that exceeds manual performance. This one-way communication of task planning, however, is limited to the constraint that the robot has to lead the user. That way, the introduction of user decisions can result in conflicts of plans with the robot which in turn can inflict frustration in users and decreases cooperative task performance. Furthermore, this concept does not go in line with the users' idea of cooperation as the robot's behaviour was sometimes hard to predict e.g. users would not know where the robot would move next. 
    
    As a starting point of addressing this problem, extended user perception was introduced to the robot which allows the estimation of the user's eye gaze in 3D space during task execution\cite{Stolzenwald:2018un}. An estimate of users' visual attention was then used to inform the robot's decisions when there was an alternative e.g. the user could prioritise subsequent goals. While this feature was preferred, particularly for temporal demanding tasks, we lag a rather sophisticated model that could be used for tasks with higher complexity. Such a model would allow the robot to infer user intention i.e. predict users' goal in the proximate future rather than reacting to immediate decisions only.  
    
    % this might be left out
    Intention inference has caught researcher's attention in recent years and promising solutions have been achieved through observing user's eye gaze \cite{Huang:2016dj}, body motion \cite{Ravichandar:2015ii} or task objects \cite{Liu:2015km}. These contributions target safe interactions between humans and stationary robots with shared workspaces. Thus, the question remains open whether there is a model which suits the setup of a handheld robot which is characterised by close shared physical dependency and a \textit{working together} rather than \textit{turn taking} cooperative strategy.
    
    Our work is guided by the following research questions
    \begin{enumerate}[label=\textbf{Q\arabic*}]
        \item How can user intention be modelled in the context of a handheld robot task?
        \label{Q1}
        \item To what extent can handheld robot user actions like picking and placing be predicted in advance?
        \label{Q2}
        \vspace{-1em}
    \end{enumerate}
    %title figure
    \begin{figure}[t]
        %\vspace{0.5em}
        \centering
        \includegraphics[width=0.99\linewidth]{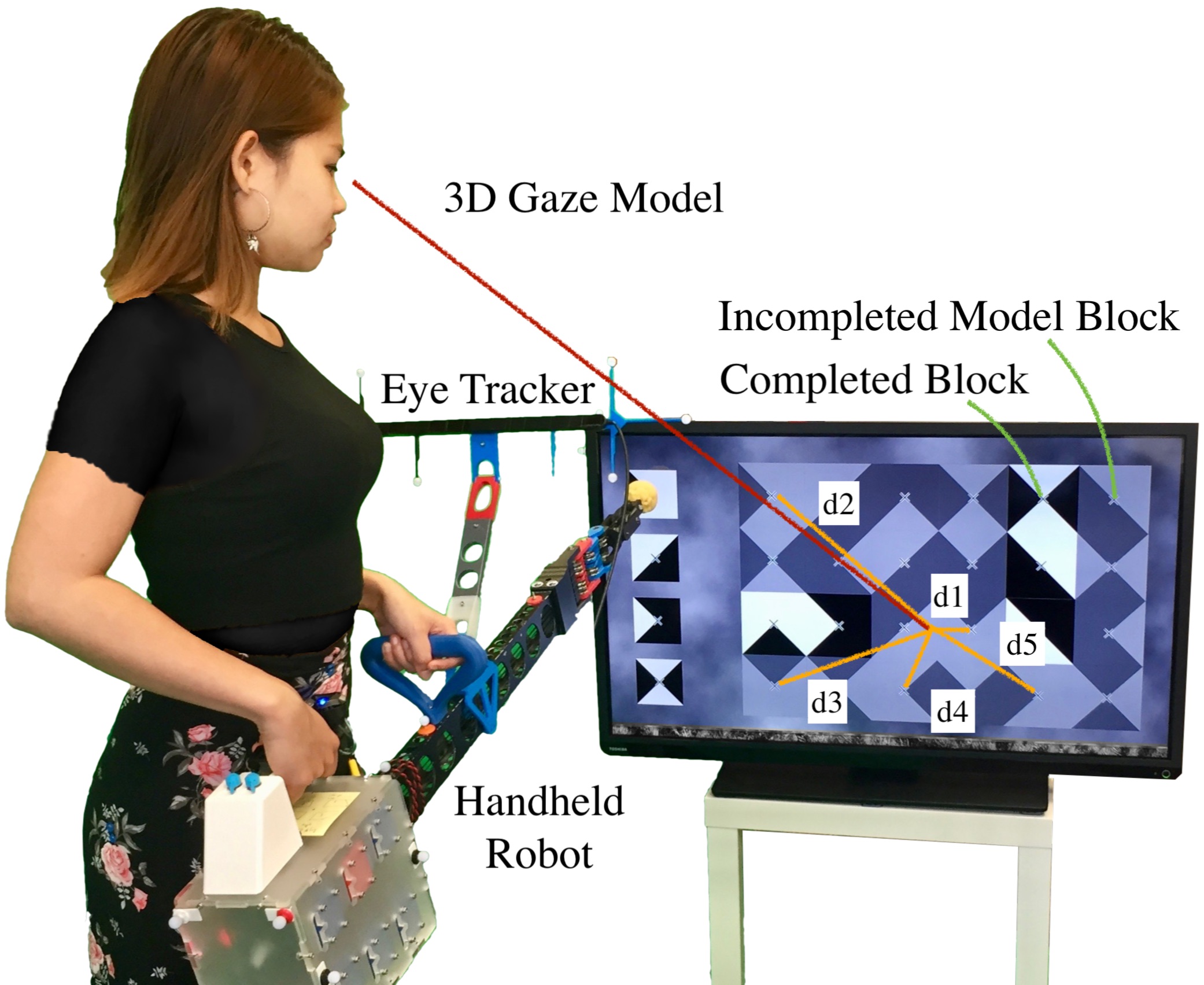}
        \caption{This picture shows a participant within our user intention prediction study. The participant uses the robot to solve an assembly task and is about to decide where to place the currently held block. Using the eye tracker, our prediction system tracks the distances $d_1$ to $d_5$ between potential candidates and the eye gaze. These are subsequently converted into visual attention profiles which are used for action prediction.}
        \label{fig:blockcopygameparticipant}
        \vspace{-1.5em}
    \end{figure}
    For our study, we use the open robotic platform\footnote{3D CAD models available from handheldrobotics.org}, introduced in \cite{GreggSmith:2016cz} in combination with an eye tracking system as reported in \cite{Stolzenwald:2018un}. Within a simulated assembly task, eye gaze information is used to predict subsequent user actions. The two principal parts of this study consist of the design of the experiment used for data collection in the first place and secondly the method of modelling user intention followed by a detailed evaluation.

    \section{Background and Related Work}
    In this section, we deliver a summary of earlier work on handheld robots and its control based on user perception. Furthermore, we review existing methods for intention inference with a focus on human gaze behaviour.
    
    \subsection{Handheld Robots}
    Early handheld robot work \cite{GreggSmith:2015bh} used a trunk-shaped robot with 4-DoF to explore issues of autonomy and task performance. This was later upgraded to a 6-DoF (joint space) mechanism \cite{GreggSmith:2016cz} and used gestures, such as pointing, to study user guidance. These earlier works demonstrate how users benefit from the robot's quick and accurate movement while the robot profits from the human's tactical motion. Most importantly, increased cooperative performance was measured with an increased level of the robot's autonomy. They furthermore found that cooperative performance significantly increases when the robot communicates its plans e.g. via a robot-mounted display \cite{GreggSmith:2016hn}. 
    
    Within this series of work, another problem was identified: the robot does not sense the user's intention and thus potential conflicts with the robot's plan remain unsolved. For example, when the user would point the robot towards a valid subsequent goal, the robot might have already chosen a different one and keep pointing towards it rather than adapting its task plan. This led to irritation and frustration in users on whom the robot's plan was imposed on. 

    Efforts towards involving user perception in the robot's task planning were made in our recent work on estimating user attention\cite{Stolzenwald:2018un}. The method was inspired by work from Land et al. who found that human's eye gaze is closely related to their manual actions \cite{Land:2016kw}. The attention model  measures the current visual attention to bias the robot's decisions. In a simulated \textit{space invader} styled task, different levels of autonomy were tested over varying configurations of speed demands. It was found that both the fully autonomous mode (robot makes every decision) and the attention driven mode (robot decides based on gaze information) outperform manual task execution. Notably, for high-speed levels, the increased performance was most evident for the attention-driven mode which was also rated more helpful and perceived rather cooperative than the fully autonomous mode. 
    
    As opposed to an intention model, the attention model would react to the current state of eye gaze information only, rather than using its history to make predictions about the user's future goals. We suggest that this would be required for cooperative task solving for complex tasks like assembly where there is an increased depth of subtasks. 
    % In an extended version, we could mention Vatsal's Arm mounted robot here e.g. \cite{Vatsal:2018wv}

    \subsection{Intention Prediction}
    Human intention estimation in the field of robotics is in part driven by the demand for safe human-robot interaction and efficient cooperation and here we review recent contributions with a broad variety of approaches. 
    
    Ravichandar et al. investigated intention inference based on human body motion. Using Microsoft Kinect motion tracking as an input for a neural network, reaching targets where successfully predicted within an anticipation time of approximately \SI{0.5}{s} prior to the hand touching the object\cite{Ravichandar:2015ii}. Similarly, Saxena et al. introduced a measure of affordance to make predictions about human actions and reached  84.1\%/74.4\% accuracy \SI{1}{s}/\SI{3}{s} in advance, respectively\cite{Koppula:2016ja}. Later, Ravichandar et al. added human eye gaze tracking to their system and used the additional data for pre-filtering to merge it with the existing motion-based model \cite{Ravichandar:2016th}. The anticipation time was increase to \SI{0.78}{s}.
    
    Huang et al. used gaze information from a head-mounted eye tracker to predict customers' choices of ingredients for sandwich making. Using a support vector machine (SVM), an accuracy of approximately 76\% was achieved with an average prediction time of \SI{1.8}{s} prior to the verbal request \cite{Huang:2015iw}.    
    In subsequent work, Huang \& Mutlu used the model as a basis for a robot's anticipatory behaviour which led to more efficient collaboration compared to following verbal commands only \cite{Huang:2016dj}.
    
    % Maybe leave out:    
    Another approach to estimating user intention is to observe task objects instead of the human which was demonstrated through 2D \cite{Liu:2014ib} and 3D \cite{Liu:2015km} block assembly tasks.
    
    We note that the above work targets intention inference purposed for \textit{external} robots which are characterised by a shared workspace with a human but can move independently. It is unclear whether these methods are suitable for close cooperation as it can be found in the handheld robot setup.

    \subsection{Human Gazing Behaviour}
    The intention model we present in this paper is mainly driven by eye gaze data. Therefore, we review work on human gaze behaviour to inform the underlying assumptions of our model. 
    
    One of the main contributions in this field is the work by Land et al. who found that fixations towards an object often precede a subsequent manual interaction by around \SI{0.6}{s} \cite{Land:2016kw}. Subsequent work revealed that the latency between eye and hand varies between different tasks \cite{Land:2001hl}. Similarly, Johansson et al. \cite{Johansson:2001ck} found that objects are most salient for human's when they are relevant for tasks planning and preceding saccades were linked to short-term memory processes in \cite{Mennie:2006fo}. 
    
    The purpose of preceding fixations in manual tasks was furthermore explored through virtual \cite{Ballard:1995iy} and real \cite{Pelz:2001fb} block design tasks. The results show that humans gather information through vision \textit{just in time} rather than memorising e.g. all object locations. 
    
\begin{figure}[h]
    \vspace{-0.5em}
    \centering
    \includegraphics[width=0.88\linewidth]{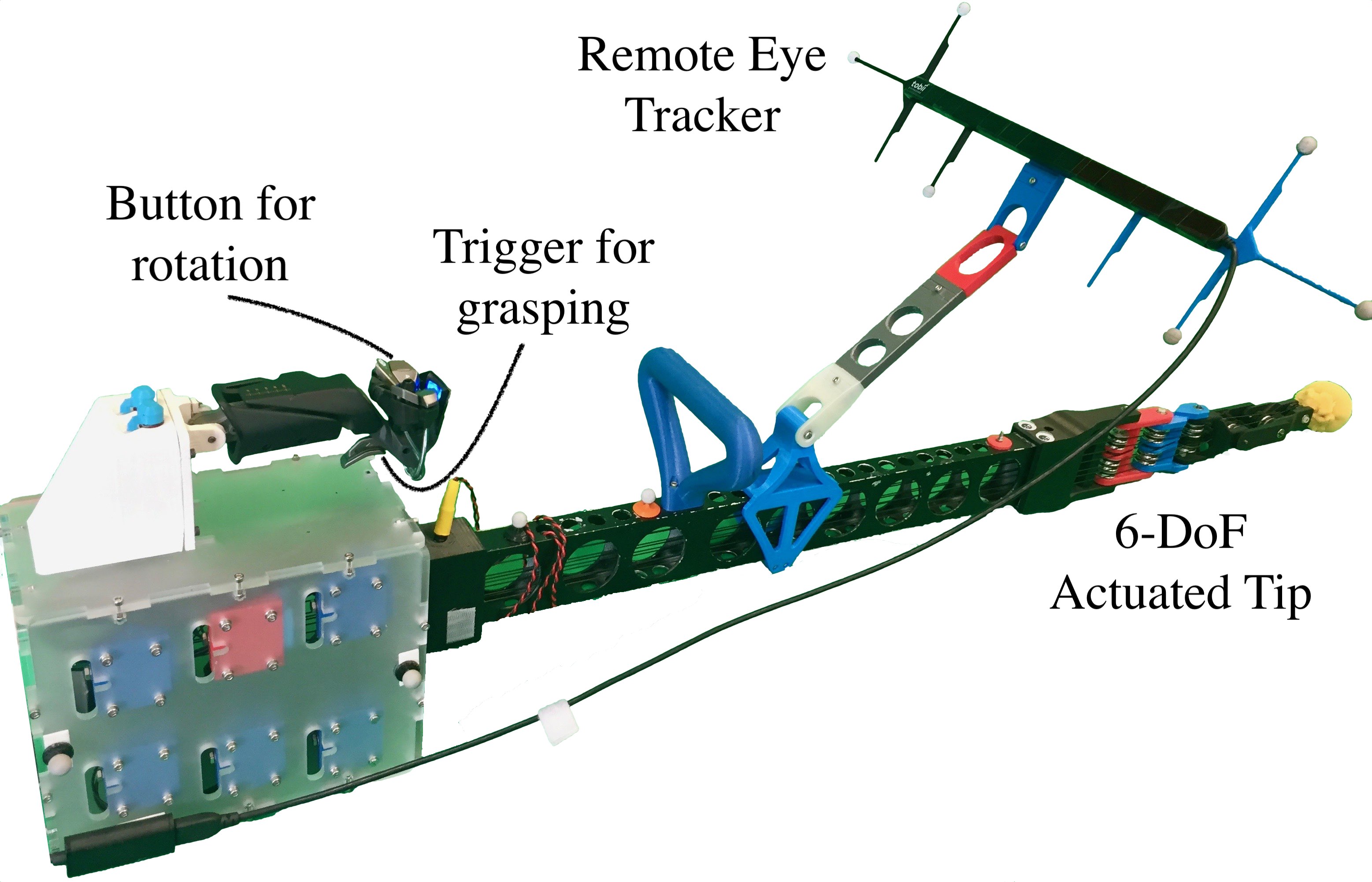}
    \caption{The Handheld robot used for our study. It features a set of input buttons and a trigger at the handle, a 6-DoF tip and user perception through gaze tracking as reported in \cite{Stolzenwald:2018un}.}
    \label{fig:frontprofilerobotlabeled}
    \vspace{-1em}
\end{figure}

\section{Prediction of User Intention}
    In this section, we describe how intention prediction is modelled for the context of a handheld robot on the basis of an assembly task. The first part is about how users' gaze behaviour is captured and quantified within an experimental study. In the second part, we describe how this data is converted into features and how these were used to predict user intent.

    \subsection{Data Collection}
\begin{figure}[h]
        \vspace{0.5em}
        \centering
        \includegraphics[width=0.88\linewidth]{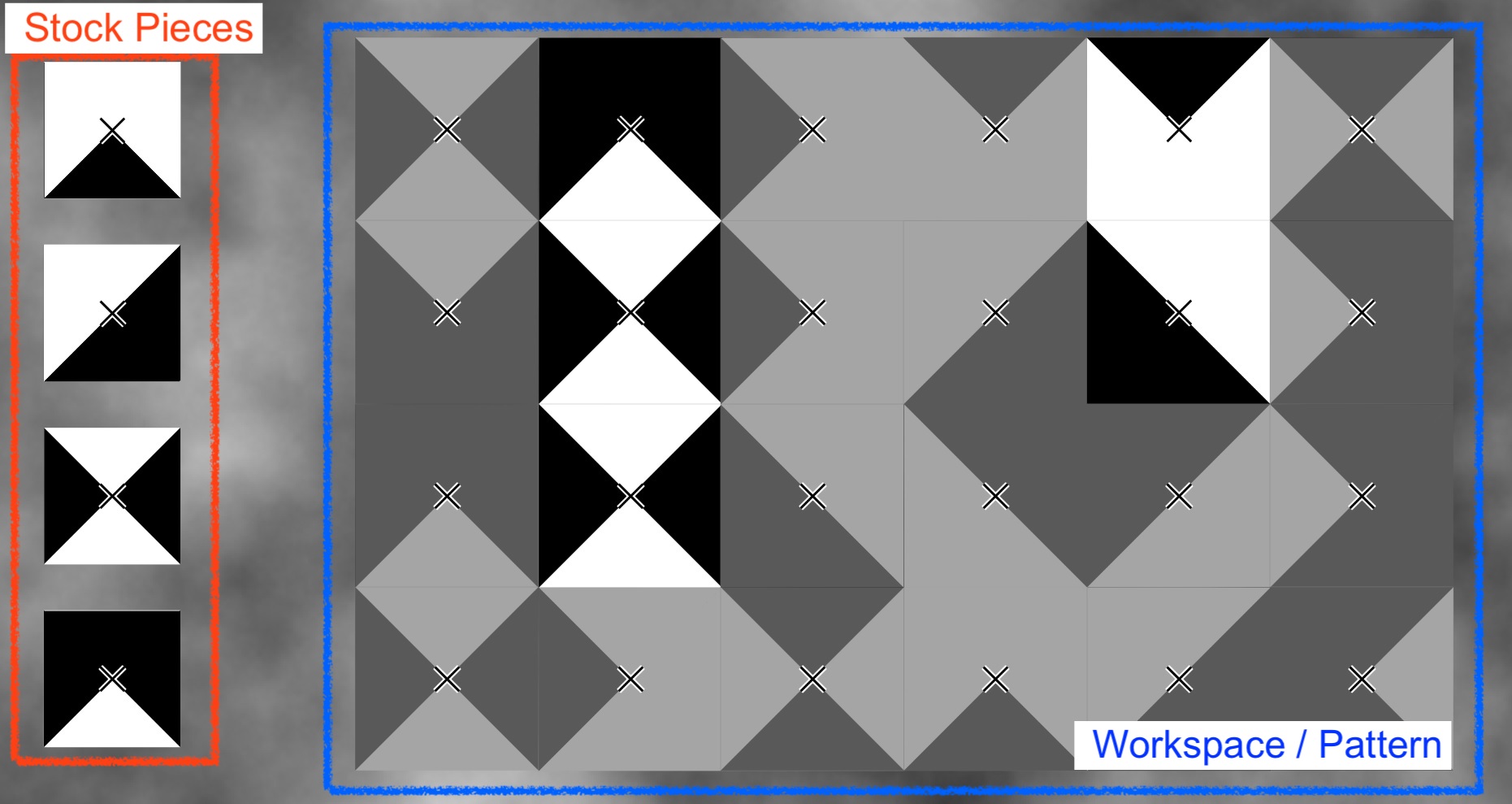}
        \caption{Layout of the block copy task on a TV display. The area is divided into stock (red) and workspace (blue). The shaded pattern pieces in the workspace area have to be completed by placing the associated pieces from the stock using the real robot.}
        \label{fig:blockcopyinitexampleareas}
        \vspace{-1em}
    \end{figure}

\begin{figure}[h]
        \centering
        \includegraphics[width=0.88\linewidth]{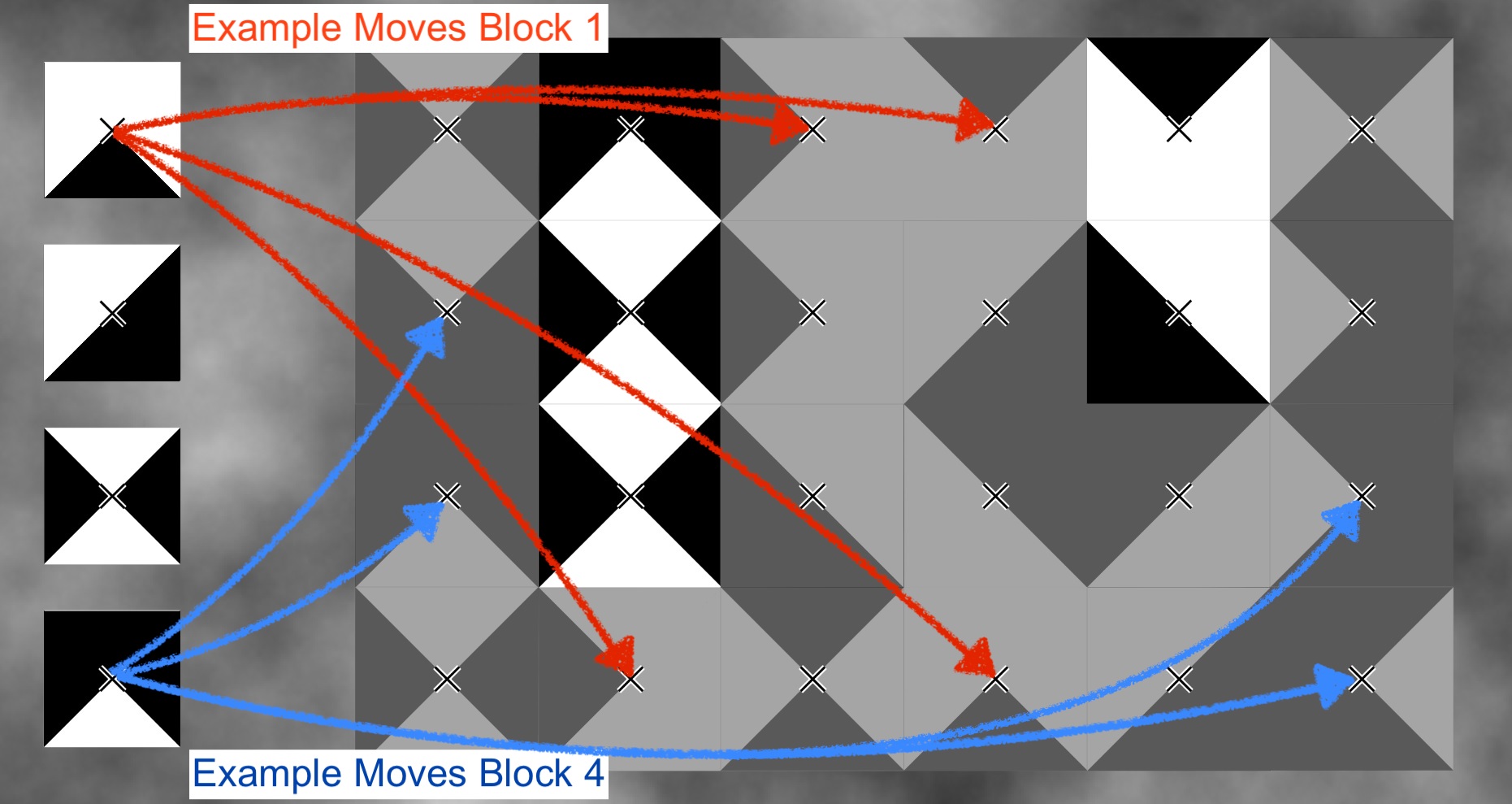}
        \caption{Examples of possible moves for block 1 and 4. A stock piece has to be moved to an associated piece in the pattern and match the model's orientation to complete it.}
        \label{fig:blockcopyinitexamplemoves}
        \vspace{-2em}
    \end{figure}

    As an example task for data collection, we chose a simulated version of a block copying task which has been used in the context of work about hand-eye motion before \cite{Ballard:1995iy,Pelz:2001fb}. Participants of the data collection trials were asked to use the handheld robot (cf. figure \ref{fig:frontprofilerobotlabeled}) to pick blocks from a stock area and place them in the workspace area at one of the associated spaces indicated by a shaded model pattern. The task was simulated on a \SI{40}{inch} LCD TV display and the robot remained motionless to avoid distraction. Rather than using coloured blocks, we drew inspiration from an block design IQ test \cite{Miller:2009ib} and decided to use ones that are distinguished by a primitive black and white pattern. That way, a match with the model would in addition depend on the block's orientation which adds some complexity to the tasks; plus, the absence of colours aims to level the challenge for people with colour blindness. An overview of the task can be seen in figure \ref{fig:blockcopyinitexampleareas}, figure \ref{fig:blockcopyinitexamplemoves} shows examples of possible picking and placing moves. 
    
    
    In order to pick or place pieces, users have to point the robot's tip towards and close to the desired location and pull/release a trigger in the handle. The position of the robot and its tip is measured via a motion tracking system\footnote{Opti Track: https://optitrack.com}\hspace{-0.4em}. The handle houses another button which users can use to turn the grabbed piece by \SI{90}{deg} for each activation. The opening or closing process of the virtual gripper takes \SI{1.3}{s} which is animated in the screen. If the participant tries to place a mismatch, the piece goes back to the stock and has to be picked up again. Participants are asked to solve the task swiftly and it is completed when all model pieces are copied. Figure \ref{fig:blockcopygameparticipant} shows an example of a participant solving the puzzle. 
    
    For the data collection, 16 participants (7 females, $m_{age}$ = 25, \textit{SD} = 4) were recruited to complete the block copy task, mostly students from different fields. Participation was on a voluntary basis and there was no financial compensation for their time, however, many considered the task as a fun game. Each completed one practice trial to get familiar with the procedure, followed by another three trials for data collection, where stock pieces and model pieces were randomised prior to execution. The pattern consists of 24 parts with an even count of the 4 types. 
    % if anything than this can go
    The task starts with 5 pre-completed pieces to increase the diversity of solving sequences leaving 19 pieces to complete by the participant. That way, a total amount of 912 episodes of picking and dropping were recorded.
    
    Throughout the task execution, we kept track of the user's eye gaze using a robot-mounted remote eye tracker in combination with a 3D gaze model as introduced in \cite{Stolzenwald:2018un}. That information was used to measure the Euclidean gaze distance $d(t)$ for each object in the scene over time $t$, that is the distance between an object's centre point and the intersection between eye gaze and screen. In the following, we call the record of $d(t_0)$ over a time $t<t_0$ for an object \textit{gaze history}. Moreover, we recorded the times for picking and placing actions for later use in auto-segmentation. 
    % extended version: add how we dealt with gaze gaps
    
    With 912 recorded episodes, 4 available stock pieces and 24 pattern parts, we collected 3648 gaze histories for stock parts prior to picking actions and 21888 for pattern pieces.

    \subsection{User Intention Model}
    
    In the context of our handheld robot task, we define intention as the user's choice of which object to interact with next i.e. which stock piece to pick and on which pattern field to place it. 
    
    Based on our literature review, our modelling is guided by the following assumptions. 
    \begin{enumerate}[label=\textbf{A\arabic*}]
        \item An intended object attracts the users' visual attention prior to interaction. \label{A1}
        \item During task planning, the users' visual attention is shared between the intended object and other (e.g. subsequent) task-relevant objects.\label{A2}
    \end{enumerate}
    
    % for an extended version we could mention that we also assume that tasks were planned in a sequence with many fixations (just in time) rather than memorising everything at the beginning.
        
    As a first step towards the feature construction, the recorded gaze history (measured as a series of distances as described above) is converted into a visual attention profile (VAP) $P_{gazed,i}(t)$ through the following equation:        
    \vspace{-0.5em}
    \begin{equation}    
    \vspace{-0.5em}
    P_{gazed,i}(t) = \exp(\frac{-d_i(t)^2}{2\sigma^2})
    \end{equation}    
    \begin{figure}[b]        
        \vspace{-1.5em}
        \centering
        \includegraphics[width=0.99\linewidth]{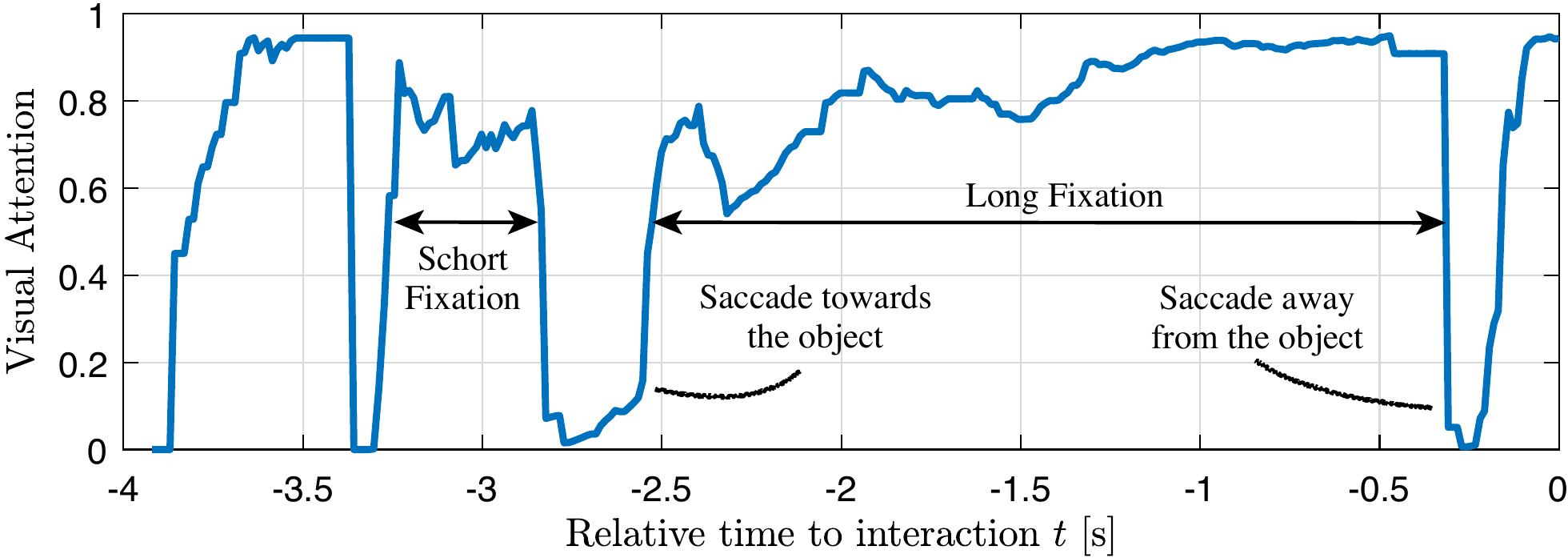}        
        \vspace{-1.5em}
        \caption{Illustration of changing visual attention over time within the anticipation window of the prediction model for an individual object.
        % extended version: note how it includes saccades and fixation durations
        }
        \label{fig:visualattentionprofile}
    \end{figure}
    where  $d_i(t)$ is the time dependent gaze distance of the i-th object (cf. figure \ref{fig:blockcopygameparticipant}). $\sigma$ defines the gaze distance resulting into a significant drop of $P_{gazed}$ and was set to \SI{60}{\mm} based on the pieces' size and tracking tolerance. Here, we define an object's gaze profile at the time $t_0$ as the collection of $P_{gazed}$ over a \SI{4}{s} anticipation window prior to $t_0$. Due to the data update frequency of \SI{75}{Hz} the profile is discretised into a vector of 300 entries. An example can be seen in figure \ref{fig:visualattentionprofile}.
    
    The prediction for picking and placing actions was modelled separately as they require different feature sets. As studies about gaze behaviour during block copying \cite{Ballard:1995iy} suggest that the eye gathers information about both what to pick and where to place it prior to picking actions, we combined pattern and stock information for picking predictions for each potential candidate to chose, resulting in the features:
    \begin{enumerate}[leftmargin=.42in]
        \item[$F_{pick,1}$] The VAP of the object itself.
        \item[$F_{pick,2}$] The VAP of the matching piece in the pattern. If there are several, the one with the maximum visual attention is picked. 
    \end{enumerate}
    This goes in line with our assumptions \ref{A1}, \ref{A2}. Both features are vectors of real numbers between 0 and 1 with a length of $n = 300$.
    
    For the prediction of the dropping location, \ref{A2} is not applicable as the episode finishes with the placing of the part. However, we hypothesise that it would be unlikely that a participant would drop the piece on a mismatch or on a match that is already completed. This results in the following feature set which is calculated for each dropping location:
    \begin{enumerate}[leftmargin=.42in]
        \item[$F_{place,1}$] The VAP of the object itself (vector, $n = 300$).
        \item[$F_{place,2}$] Whether or not the location matches with what is hold by the gripper (boolean).
        \item[$F_{place,3}$] Whether or not this pattern space is already completed (boolean).
    \end{enumerate}

    As prediction models for picking and placing intention, we used SVMs \cite{Hearst:1998ew} as this type of supervised machine learning model was used for similar classification problems in the past, e.g. \cite{Huang:2015iw}. We divided the sets of VAPs into two categories, one where the associated object was the intended object (labelled as \texttt{chosen = 1}) and another one for the objects that were not chosen for interaction (labelled as \texttt{chosen = 0}). Training and validation of the models were done through 5-fold cross validation \cite{Kohavi:1995wf}. 
    
    The accuracy of predicting the \texttt{chosen} label for individual objects is 89.6\% for picking actions and 98.3\% for placing intent. However, sometimes the combined decision is conflicting e.g when several stock pieces are predicted to be the intended ones. This is resolved by selecting the one with the highest probability $P($\texttt{chosen }$ = 1)$ in a one-vs-all setup \cite{Rifkin:2004vf}. This configuration was tested for scenarios with the biggest choice e.g. when all 4 stock parts (random chance = 25\%) would be a reasonable choice to pick or when the piece to be placed matches 4 to 6 different pattern pieces (random chance = 17-25\%). This results in a correct prediction rate of 87.9\% for picking and 93.25\% for placing actions when the VAPs of the time up to just before the action time is used.     

    \section{Results}
\begin{figure*}[t!]
        \vspace{0.5em}
        \centering
        \begin{subfigure}[t]{0.48\textwidth}
            %put what is normally in figure here
            \centering
            \includegraphics[width=0.99\linewidth]{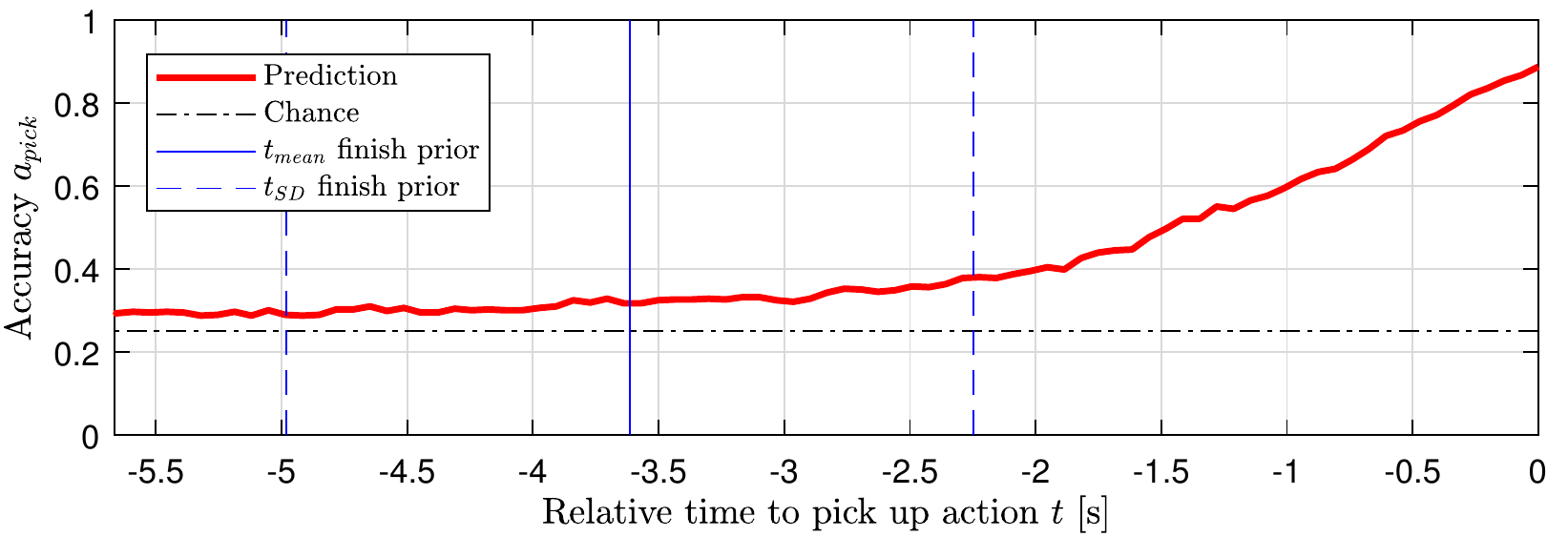}
            \caption{\scriptsize performance of intention prediction for the user's pick up action with respect to the time $t$ prior to grasping the object from the stock.
                % This includes 892 picking actions of 15 participants (1 discarded) hence 3568 data points of pick and 21,408 data points for placing.  If we use visual attention only for prediction, we get up to 77.59\% accuracy.
            }
            \label{fig:pickuppredictionovertime}
        \end{subfigure}%
        ~ 
        \begin{subfigure}[t]{0.48\textwidth}
            \centering
            \includegraphics[width=0.99\linewidth]{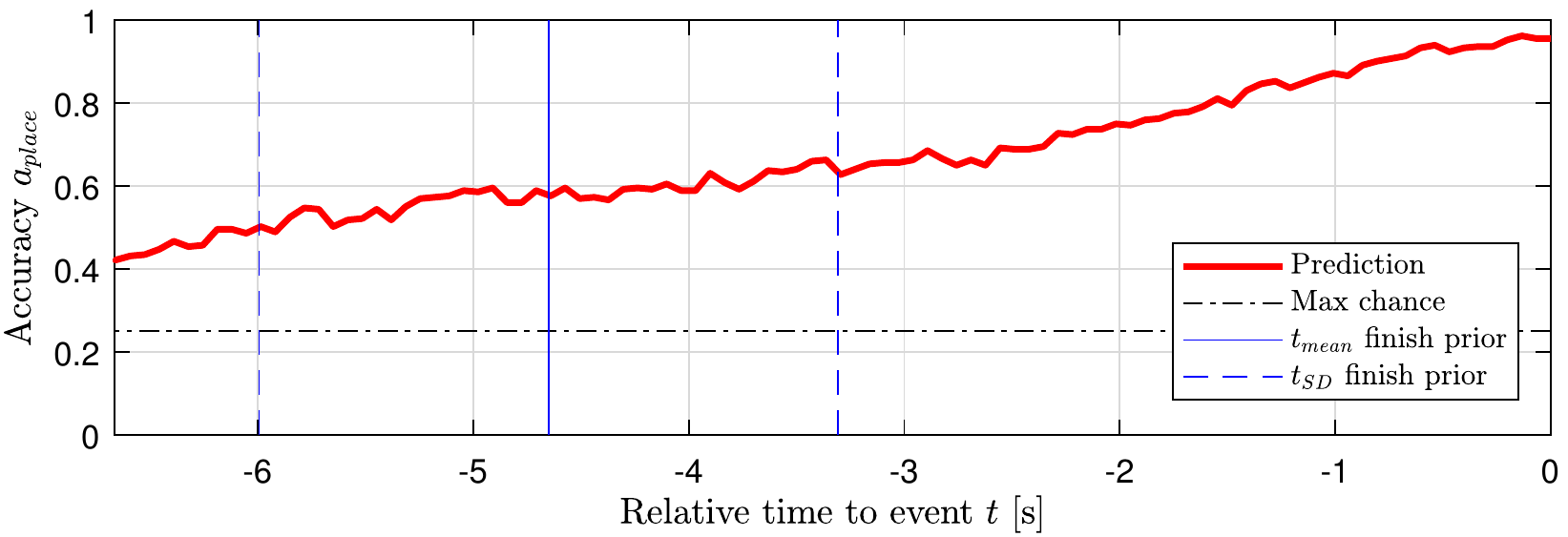}
            \caption{\scriptsize Performance of intention prediction for the user's dropping action with respect to the time $t$ prior to placing the part in the pattern. 
                % Chance = 25 to 17\%. If we use visual attention only for prediction, we get up to 81.67\% accuracy.
            }
            \label{fig:placepredictionovertime}
        \end{subfigure}
        \caption{These diagrams show how the prediction accuracies of our presented models change over the time relative to the action to predict. (a) shows the performance of pick up predictions, that is the success rate of predicting which one out of four pieces (chance = 25\%) the user will pick up next. Similarly, (b) shows the performance of predicting where the user will place the picked piece. Depending on the number of available matching pattern locations (between 4 and 6), the chance of choosing correct randomly is up to 25\%. Note that the pick up prediction model has a high steep slope short before the action event, whereas the dropping action can be predicted reliably short after the mean pick up time $t_{prior}$.}
        \label{fig:predictionOverTime}    
        \vspace{-0.8em}
    \end{figure*}

    Having trained and validated the intention prediction model for the case where VAPs range from -4 to 0 seconds prior to the interaction with the associated object, we are now interested in knowing to what extent the intention model predicts accurately at some time $t_{prior}$ prior to interaction. To answer this question, we extend our model analysis by calculating a $t_{prior}$-dependent prediction accuracy. Within a 5-fold cross validation setup, the \SI{4}{s}-anticipation window is iteratively moved away from the time of interaction and the associated VAPs are used to make a prediction about the subsequent user action using the trained SVM models. The validation is based on the formentioned low-chance subsets, so that the chance of correct prediction through randomly selecting a piece would be $\leq25\%$. The shift of the anticipation window over the data set is done with a step width of 1 frame (\SI{13}{ms}). This is done for both the case of predicting which piece is picked up next as well as inferring intention concerning where it is going to be placed. For the time offsets $t_{prior}$ = 0,0.5 and 1 seconds, the prediction of picking actions yields an accuracy $a_{pick}$ of 87.94\%, 72.36\% and 58.07\%. The performance of the placing intention model maintains a high accuracy over a time span of \SI{3}{s} with an accuracy $a_{place}$ of 93.25\%, 80.06\% and 63.99\% for the times $t_{prior}$ = 0,1.5 and 3 seconds. In order to interpret these differences in performance, we investigated whether there is a difference between the mean duration of picking and placing actions. We applied a two-sample t-test and found that the picking time (mean = \SI{3.61}{s}, \textit{SD} = \SI{1.36}{s}) is significantly smaller than the placing time (mean = \SI{4.65}{s}, \textit{SD} = \SI{1.34}{s}), with $ p < 0.001, t = -16.12$. A detailed profile of the time-dependent prediction performance for each model can be seen in figure \ref{fig:predictionOverTime}.

    \subsection{Qualitative Analysis}
    For an in-depth understanding of how the user intention models respond to different gaze patterns, we investigate the prediction profile i.e. the change of the prediction over time, for a set of typical scenarios.\\
    \subsubsection{One Dominant Type}%\hspace*{\fill} \\
\begin{figure*}[t!]
        \centering
        \begin{subfigure}[t]{0.32\textwidth}
            \centering
            \includegraphics[width=0.99\linewidth]{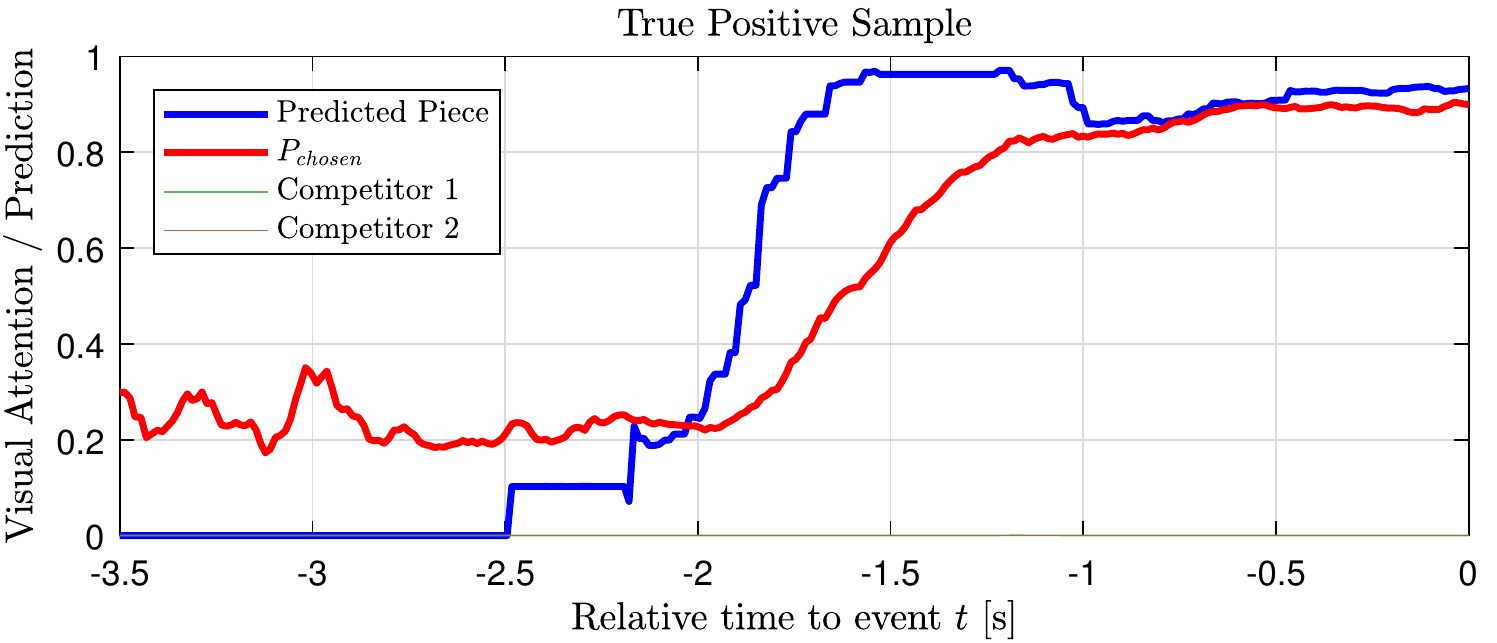}
            \caption{\scriptsize One piece receives most of the user's visual attention prior to placing}
            \label{fig:placingcorrecttruepositive1dominant3}
        \end{subfigure}
        ~ 
        \begin{subfigure}[t]{0.32\textwidth}
            \centering
            \includegraphics[width=0.99\linewidth]{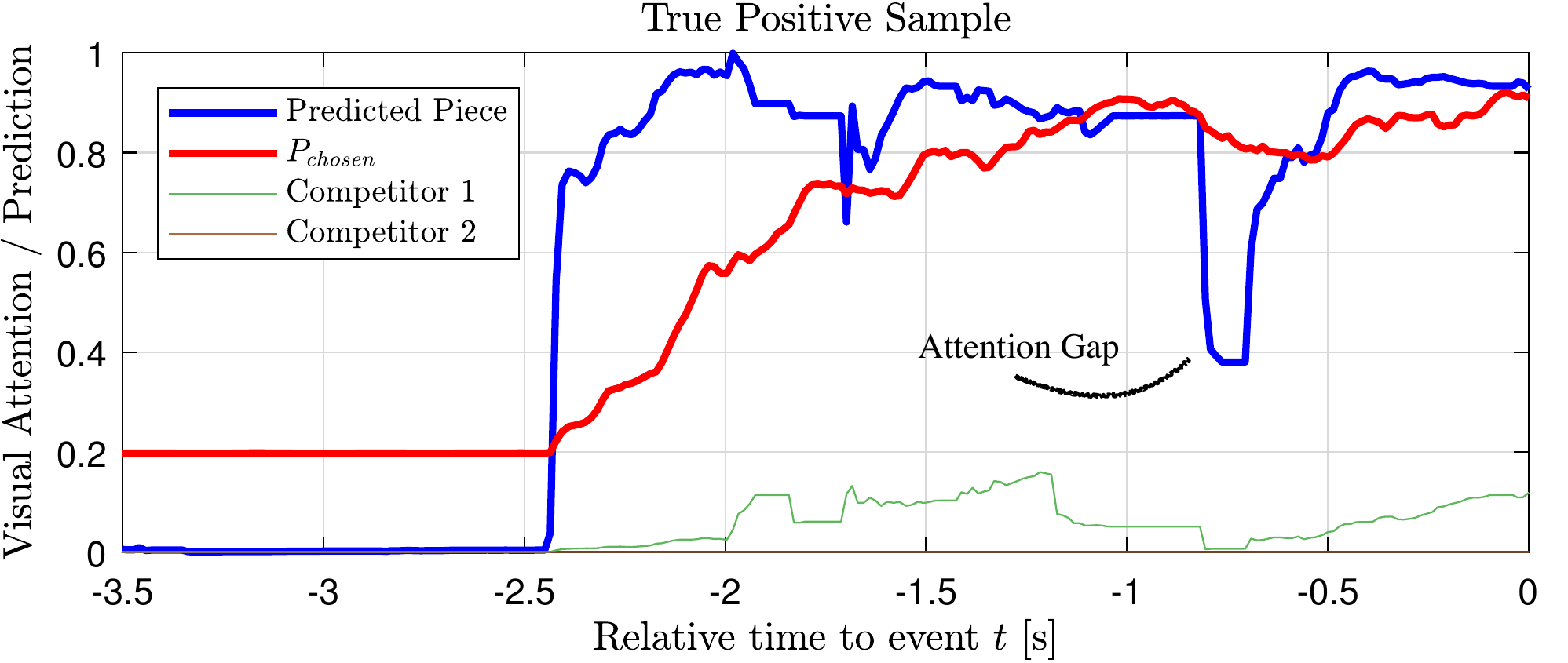}
            \caption{\scriptsize Dominance of visual attention with break gap: the prediction model maintains the prediction}
            
            \label{fig:placingcorrecttruepositive1dominant2}
        \end{subfigure}
        ~ 
        \begin{subfigure}[t]{0.32\textwidth}
            \centering
            \includegraphics[width=0.99\linewidth]{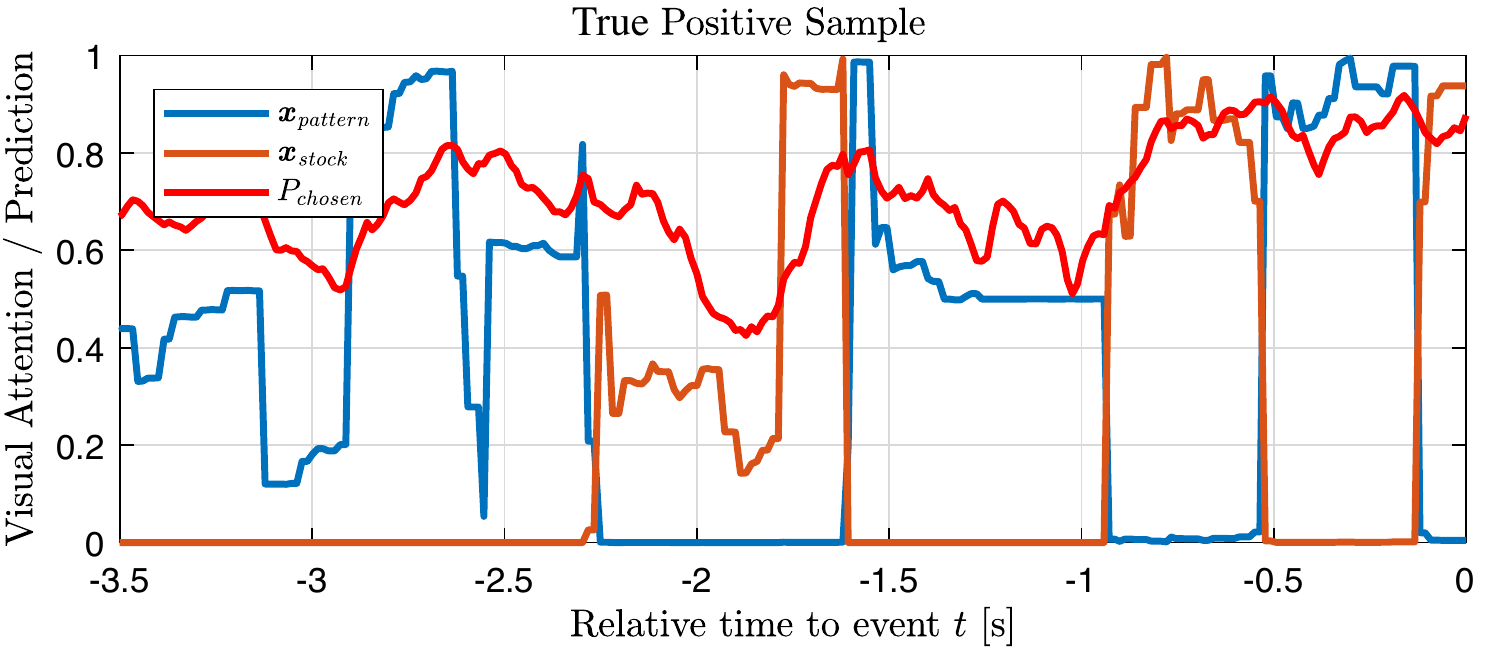}
            \caption{\scriptsize One type dominant: User gaze alters between stock piece and matching workspace location}
            \label{fig:pickupcorrecttuenegativethereandbackwideformat}
        \end{subfigure}%
    
%        % this is for placing only
%        \begin{subfigure}[t]{0.32\textwidth}
%            \centering
%            \includegraphics[width=0.99\linewidth]{Pictures/Placing_Correct_True_Positive_1Dominant_1}
%            \caption{\scriptsize  Long duration of dominance in visual attention}
%            \label{fig:placingcorrecttruepositive1dominant1}
%        \end{subfigure}%
        \caption{Examples for correct predictions of \textit{one type dominant} samples. They demonstrate how long fixations result into a high probability (a), that the model is robust against short durations of absence of user attention (b) and how pattern glances and stock fixations are combined for predicting future interaction (c). }
        \label{fig:one dominant}
        \vspace{-0.8em}
    \end{figure*}

    A common observation was that the target object perceived most of the user's visual attention prior to interaction which goes in line with our assumption \ref{A1}. We call this pattern \textit{one type dominant} and a set of examples can be seen in Figure \ref{fig:one dominant}. A subset of this category is the case where the user's eye gaze alters between the piece to pick and the matching place in the pattern i.e. where to put it (cf. figure \ref{fig:pickupcorrecttuenegativethereandbackwideformat}), which supports our assumption \ref{A2}. 
    % might be left out
    Furthermore, we note that the prediction remains stable fo for the event of a short break of visual attention i.e. the user glances away and back to the same object (cf. figure \ref{fig:placingcorrecttruepositive1dominant2}). This is a contrast to an intention inference based on the last state of visual attention only, which would result in an immediate change of the prediction. 
    For the majority of these one type dominant samples both the picking and placing prediction models predict accurately.     \\    
    \subsubsection{Trending Choice}%\hspace*{\fill} \\
    While the anticipation time of the pick up prediction model lies within a second and is thus rather reactive, the placing intention model is characterised by an increase of prediction during the task i.e. low-pass characteristic. Figure \ref{fig:trending choice} shows examples with different fixation durations (figure \ref{fig:placingcorrecttruepositivetrendingchoice1},\ref{fig:placingcorrecttruepositivetrendingchoice2}) and how the user's gaze alters between two competing places (figure \ref{fig:placingcorrecttruepositivetrendingchoice3}). The prediction model is robust for these cases, however, the anticipation time is reduced in comparison to the one type dominant samples. 
    \vspace{0.5em}
    \subsubsection{Incorrect Predictions}%\hspace*{\fill} \\
\begin{figure*}[h]
        \centering
        \begin{subfigure}[h]{0.32\textwidth}
            \centering
            \includegraphics[width=0.99\linewidth]{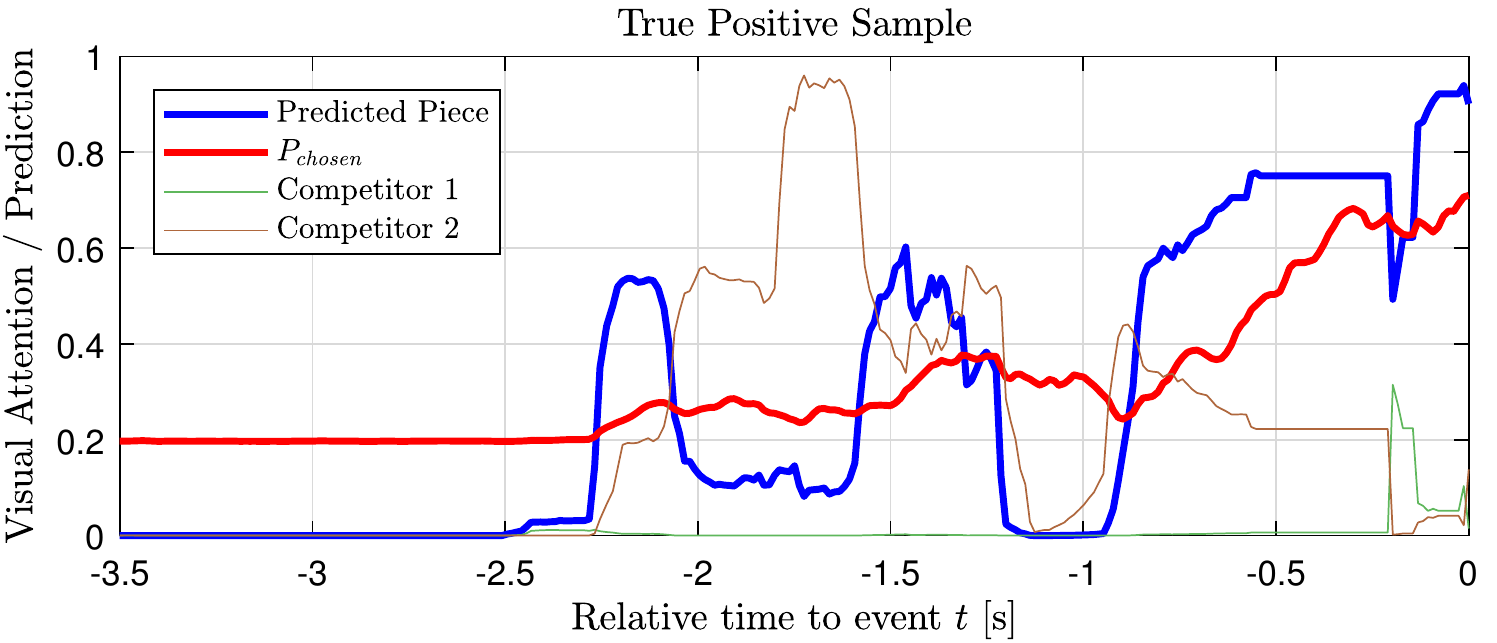}
            \caption{\scriptsize Increasing visual attention over time}
            \label{fig:placingcorrecttruepositivetrendingchoice1}
        \end{subfigure}%
        ~ 
        \begin{subfigure}[h]{0.32\textwidth}
            \centering
            \includegraphics[width=0.99\linewidth]{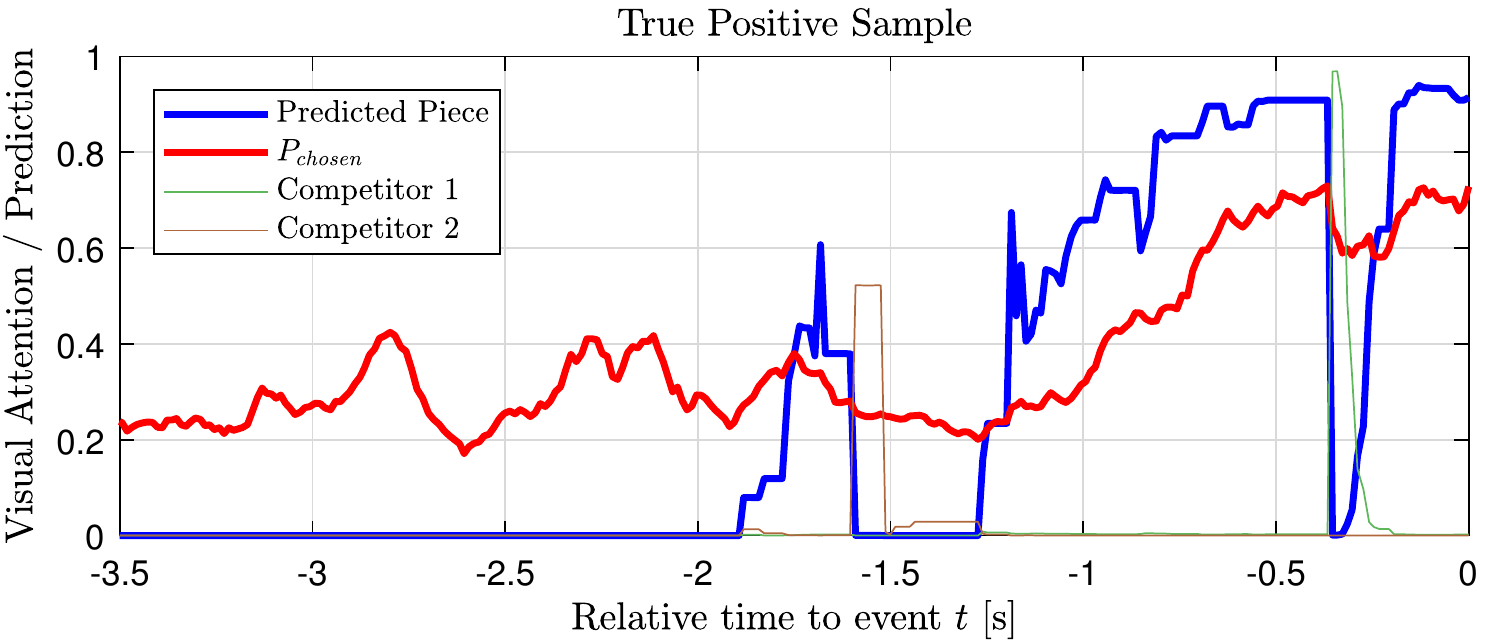}
            \caption{\scriptsize Increasing attention with competitor glances }
            \label{fig:placingcorrecttruepositivetrendingchoice2}
        \end{subfigure}
        ~ 
        \begin{subfigure}[h]{0.32\textwidth}
            \centering
            \includegraphics[width=0.99\linewidth]{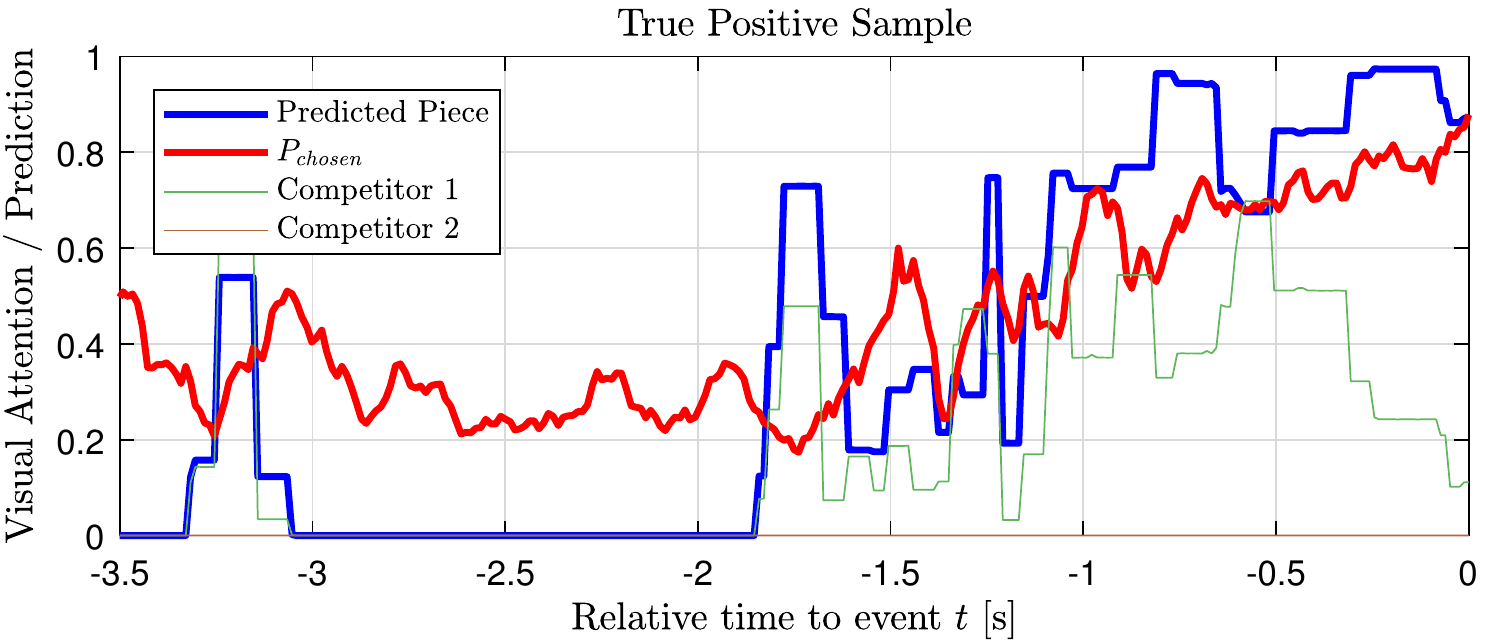}
            \caption{\scriptsize Alternating attention between competitors}
            \label{fig:placingcorrecttruepositivetrendingchoice3}
        \end{subfigure}
    
        \caption{Illustration of trending probability over a sequence of fixations (a,b) or during the selection process while choosing between competing candidates (c). }
        \label{fig:trending choice}
        \vspace{-0.8em}
    \end{figure*}

\begin{figure*}[h]
        \centering
        \begin{subfigure}[t]{0.32\textwidth}
            \centering
            \includegraphics[width=0.99\linewidth]{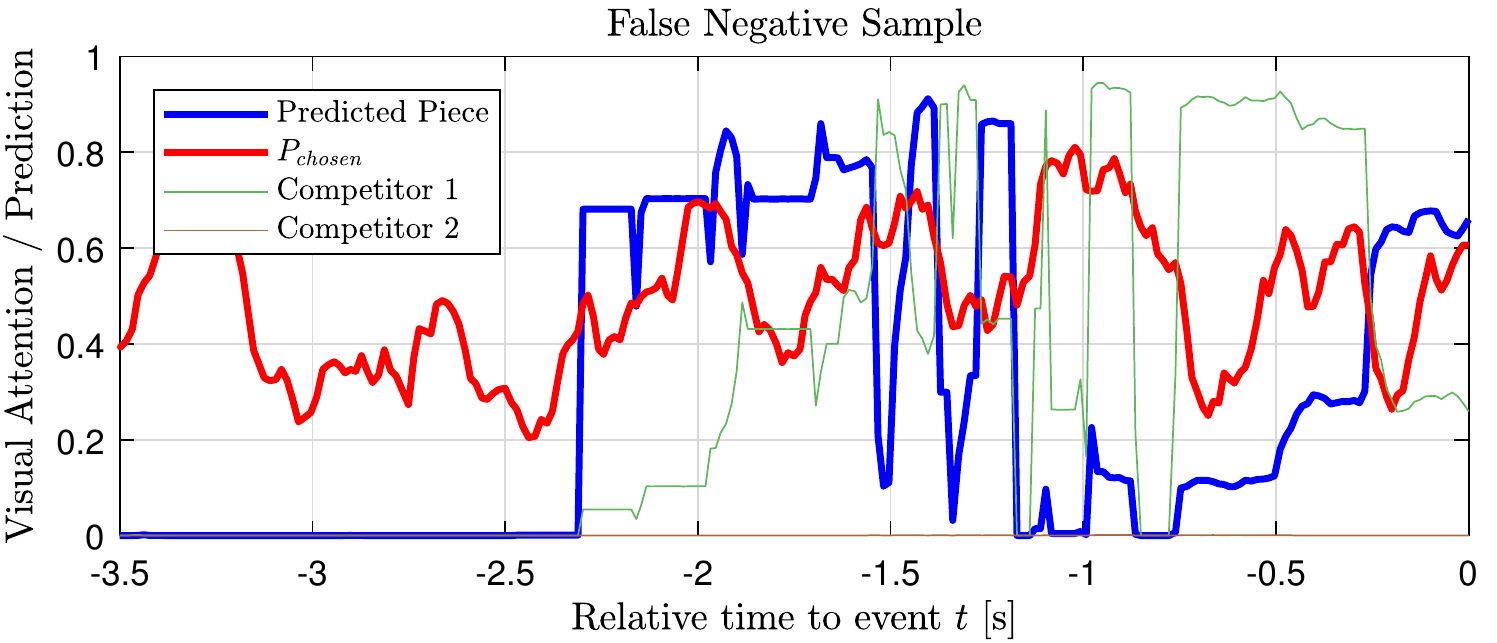}
            \caption{\scriptsize Favoured competing choice }
            \label{fig:placinginorrectfalsenegativefavourcompetingchoice2}
        \end{subfigure}%
        ~ 
        \begin{subfigure}[t]{0.32\textwidth}
            \centering
            \includegraphics[width=0.99\linewidth]{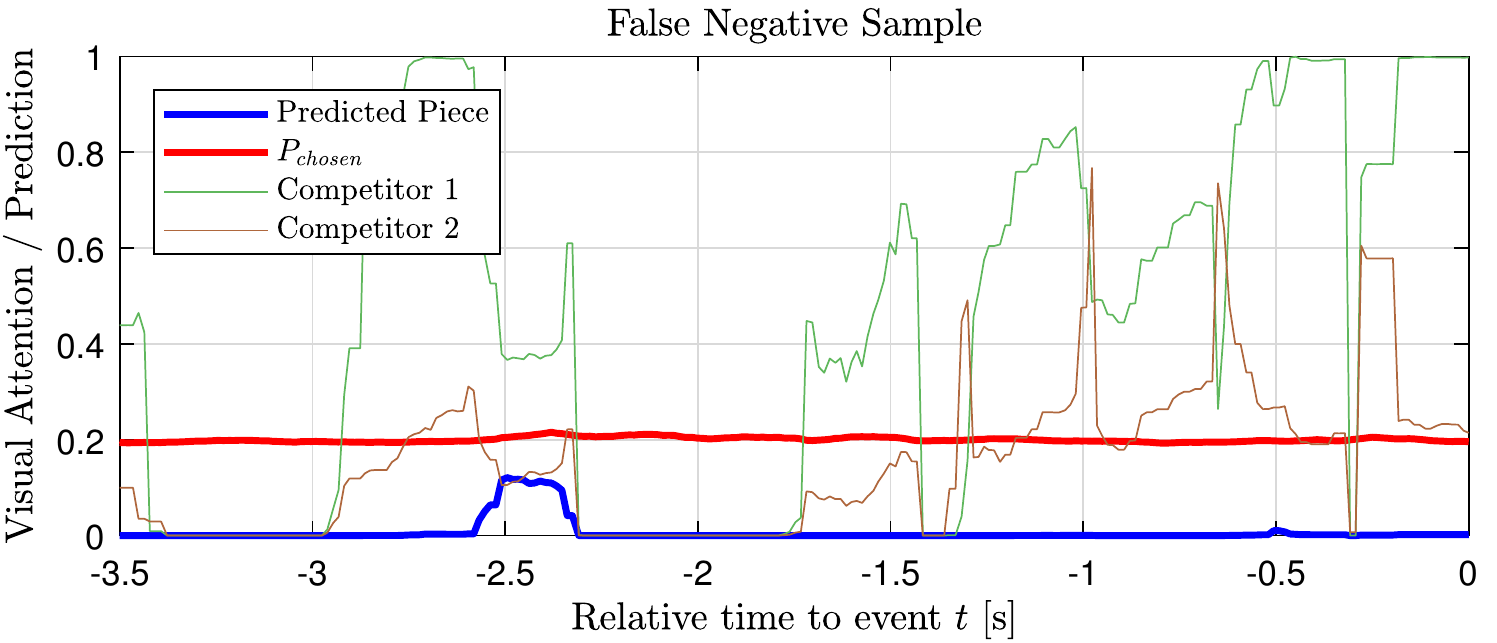}
            \caption{\scriptsize No visual intention perceived}
            \label{fig:placinginorrectfalsenegativenointendedglances}
        \end{subfigure}
        ~ 
        \begin{subfigure}[t]{0.32\textwidth}
            \centering
            \includegraphics[width=0.99\linewidth]{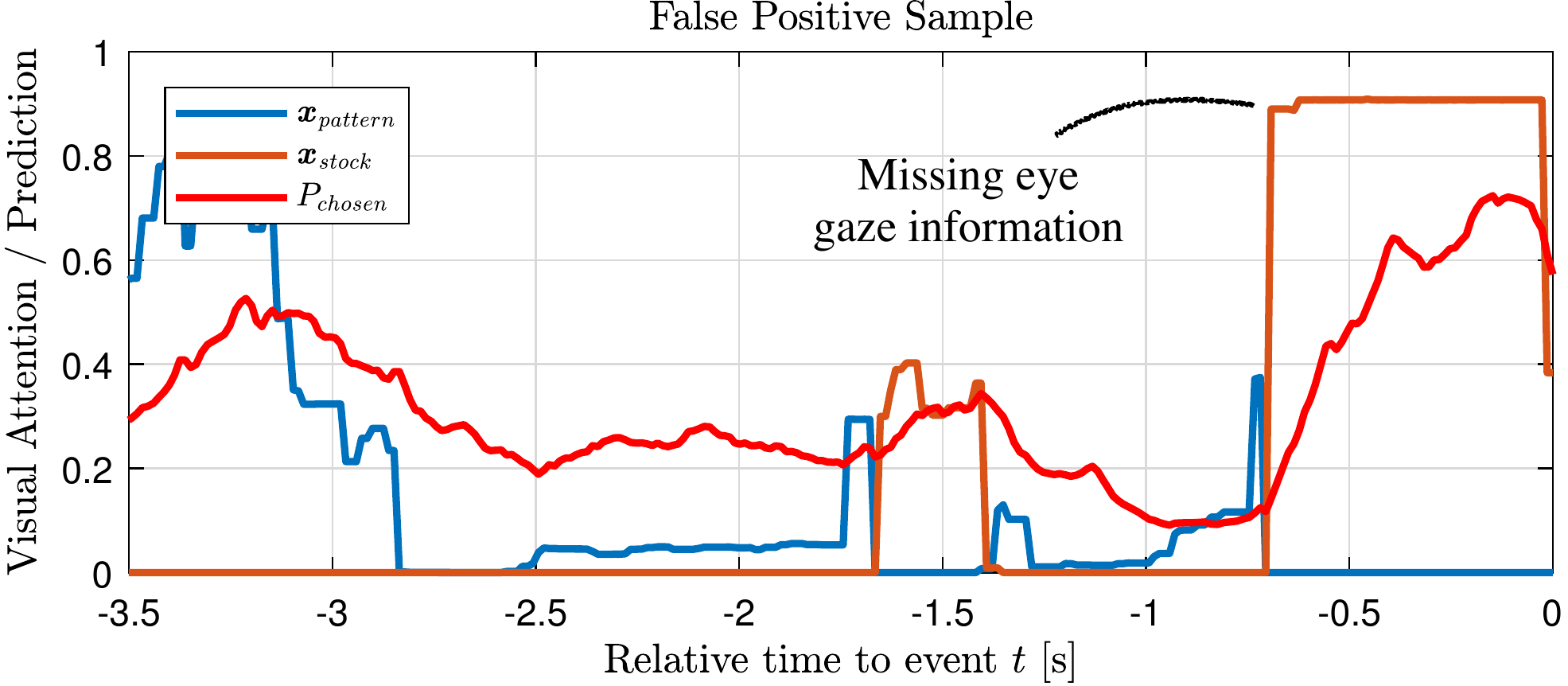}
            \caption{\scriptsize Overestimation of visual attention due to missing eye tracking information }
            \label{fig:pickupincorrectfalsepositivetrackermalfunktionwideformat}
        \end{subfigure}
        
%            % this is for placing only
%            \begin{subfigure}[t]{0.32\textwidth}
%                \centering
%  \includegraphics[width=0.99\linewidth]{Pictures/Placing_Inorrect_False_Negative_EyeGapError}
%                \caption{\scriptsize Eye tracking error (no gaze estimation)}
%                \label{fig:placinginorrectfalsenegativeeyegaperror}
%            \end{subfigure}
    \caption{Examples for incorrect predictions due to close competitors (a), the lag of fixations (b) or faulty gaze tracking (c).}
    \label{fig:Incorrect}
    \vspace{-0.5em}
    \noindent\makebox[\linewidth]{\rule{\textwidth}{0.4pt}}
    \vspace{-2.6em}
    \end{figure*}

    A common reason for an incorrect prediction is that a close competitor is chosen, for example when the user's gaze goes there and back between two potential placing candidates (figure \ref{fig:placinginorrectfalsenegativefavourcompetingchoice2}) and the incorrect choice is favoured by the model. In some rare cases there were no intended fixations recorded for the candidate prior to the interaction (cf. figure \ref{fig:placinginorrectfalsenegativenointendedglances}). In other few samples that led to faulty predictions, the eye tracker could not recognise the eyes e.g. when the robot is held so that the head was outside the trackable volume or outside the head angle range. In that case, the tracking system is unable to update the gaze model which led to over/underestimation of perceived visual attention as it can be seen in  \ref{fig:placingcorrecttruepositivetrendingchoice3}. 
    \vspace{-0.5em}

    % Prediction over time    

    % Onte Type Dominant

% Trending Choice    

    % Incorrect Samples

        % extended version: show tracker offset
%        
%        \begin{subfigure}[t]{0.32\textwidth}
%            \centering
%            \includegraphics[width=0.99\linewidth]{Pictures/Placing_Inorrect_False_Negative_OffsetError}
%            \caption{Tracker offset error.}
%            \label{fig:placinginorrectfalsenegativeoffseterror}
%        \end{subfigure}

%    
    
    % in a super extended version we would do the qualitative analysis separately for picking and placing most importantly showing how people would look there and back between workspace and stock piece
    \section{Discussion}
    In addressing research question \ref{Q1}, we proposed a user intention model based on gaze cues for the prediction of actions within a pick and place task user study as an example for handheld robot interaction. 3D user gaze was used to quantify visual attention for task-relevant objects in the scene. The derived profiles of visual intention were used as features for SVMs to predict which object will be picked up next and where it will be placed with an accuracy of 87.94 and 93.25 percent, respectively. 
    
    The prediction performance was furthermore investigated with respect to the time distance prior to the time of action to answer the research question \ref{Q2}. The proposed model allows predictions \SI{500}{ms} prior to picking actions (71.6\% accuracy) and \SI{1500}{ms} prior to dropping actions (80.06\%) accuracy. 
    
    A qualitative analysis was conducted which shows that the prediction model performs robustly for long gaze fixations on the intended object as well as for the case where users divide their attention between the intended object and related ones. Furthermore, the analysis shows the growth of the model's confidence about the prediction while the user's decision process unfolds as indicated by glances among a set of competing candidates to choose from. 
    
    We showed that, within this task, the prediction of different actions has different anticipation times i.e. dropping targets are identified quicker than picking targets. This can partially be explained by the fact that picking episodes are shorter than placing episodes. But more importantly, we observed that users planned the entire work cycle rather than planning picking and placing actions separately. This becomes evident through the qualitative analysis which shows altering fixations between the picking targets and where to place it. That way, the placing prediction model is already able to gather information at the time of picking.

    \section{Conclusion}
    Within this work, we investigated the use of gaze information to infer user intention within the context of a handheld robot scenario. A pick and place task was used to collect gaze data as a basis for an SVM-based prediction model. The results show that, depending on the anticipation time, picking actions can be predicted with up to 87.94\% accuracy and dropping actions with an accuracy of 93.25\%. Furthermore, the model allows action anticipation \SI{500}{ms} prior to picking and \SI{1500}{ms} prior to dropping. 
    
    The proposed model could be used in online anticipation scenarios to infer user intention in real time for more fluid human-robot collaboration, particularly, for the case where objects can be related to a task sequence e.g. pick and place or assembling.
    %Maybe leave out:
    %Future work will concern the integration of an anticipatory system for user actions into the context of handheld robots.
    
    {\bf Acknowledgement} 
    This work was partially supported by the German Academic Scholarship Foundation and by the UK's Engineering and Physical Sciences Research Council. Opinions are the ones of the authors and not of the funding organisations.
    
    %%%%%%%%%%%%%%%%%%%%%%% figure template
%    %whole page figure
%    \begin{figure*}[t!]
%        \centering
%        \begin{subfigure}[t]{0.32\textwidth}
%            %put what is normally in figure here
%        \end{subfigure}%
%        ~ 
%        \begin{subfigure}[t]{0.32\textwidth}
%        \end{subfigure}
%        ~ 
%        \begin{subfigure}[t]{0.32\textwidth}
%        \end{subfigure}
%        \caption{My Caption}
%    \end{figure*}
%    
    
    %%%%%%%%%%%%%%%%%%%%%%%%       BIBLIOGRAPHY
    \vspace{6em}
    {\color{white}.} %hack bib to the right place
    \bibliographystyle{unsrt} %numbers
    \bibliography{references}

    
    %\vfill
    

\end{document}